\documentclass[preprint,12pt]{elsarticle}

\usepackage{amssymb}
\usepackage{amsmath}

\journal{Engineering Applications of Artificial Intelligence}

\begin{document}

\begin{frontmatter}



\title{A Systematic Study of Compression Ordering for Large Language Models}


\author[label1]{Shivansh Chhawri}
\author[label1]{Rahul Mahadik}
\author[label1]{Suparna Rooj}
\affiliation[label1]{equal contribution}


\begin{abstract}
Large Language Models (LLMs) require substantial computational resources, making model compression essential for efficient deployment in constrained environments. Among the dominant compression techniques—knowledge distillation, structured pruning, and low-bit quantization—their individual effects are well studied, but their interactions and optimal sequencing remain unclear. This work systematically examines how these techniques perform both independently and in combination when applied to the Qwen2.5-3B model. We evaluate multiple compression pipelines, including single, and proposed three-technique sequences, using perplexity, G-Eval, clarity, prompt alignment, and compression ratio as metrics. Our experiments show that quantization provides the greatest standalone compression, while pruning introduces moderate quality degradation. Critically, the ordering of techniques significantly affects the final model quality: the sequence Pruning → Knowledge Distillation → Quantization (P-KD-Q) yields the best balance, achieving a $3.68\times$ compression ratio while preserving strong instruction-following and language understanding capabilities. Conversely, pipelines applying quantization early suffer severe performance degradation due to irreversible information loss that impairs subsequent training. Overall, this study offers practical insight into designing effective, ordering-aware compression pipelines for deploying LLMs in resource-limited settings.
\end{abstract}




\end{frontmatter}



\section{Introduction}
\label{sec:intro}
Over the past decade, artificial intelligence has experienced an extraordinary surge in capability and adoption, evolving from traditional machine learning methods to powerful deep learning systems and, more recently, to advanced generative models. This rapid growth has been fueled by breakthroughs in neural network architectures, large-scale computing, and the availability of massive datasets. Today, AI is profoundly transforming diverse domains—including healthcare \cite{shaheen2021applications}, finance  \cite{cao2022ai}, manufacturing \cite{plathottam2023review}, defense, etc. Despite remarkable progress in AI, machines do not inherently understand or generate human language without sophisticated computational models. Language modeling \cite{chowdhary2020natural}, a core task in natural language processing, focuses on predicting the next token—word or character—within a sequence, enabling systems to capture linguistic structure and contextual meaning which before the emergence of AI was a critical bottleneck \cite{jozefowicz2016exploring}. Large language models (LLMs) have rapidly grown in popularity due to their ability to perform a wide range of tasks such as text generation, reasoning, summarization after the publication of the attention based transformer model \cite{vaswani2017attention} research paper, which is the base of all the LLMs. The emergence of LLMs, image generation systems, and diffusion-based techniques highlights the accelerating pace of innovation, marking a significant shift in the way intelligent systems are developed and deployed.

Research on large language models has advanced rapidly since the introduction of transformer-based architectures, beginning with OpenAI’s Generative Pre-trained Transformer (GPT) series, where GPT \cite{radford2018improving}, GPT-2 \cite{radford2019language} and GPT-3 \cite{brown2020language} demonstrated that scaling the size of the model and training data significantly improve language understanding and generation. This paradigm was further expanded by Google’s BERT \cite{devlin2019bert} and T5 \cite{raffel2020exploring}, which highlighted the effectiveness of bidirectional and sequence-to-sequence pretraining strategies. Meta’s LLaMA models \cite{touvron2023llama} introduced a family of efficient high-performance LLMs trained on carefully curated datasets, enabling strong results with fewer parameters. Concurrently, Anthropic’s Claude models \cite{bai2022constitutional} emphasized constitutional AI and safer alignment techniques for large-scale language systems. More recently, models such as PaLM \cite{chowdhery2023palm} and Qwen \cite{bai2023qwen} have continued to push the boundaries of multilingual reasoning, instruction following, and generalization. Together, these works form a rich body of literature that charts the evolution of LLMs from early transformer models to today’s powerful foundation models used across research and industry.

Although large language models have demonstrated remarkable performance across diverse tasks, their increasing scale brings significant challenges that limit their practical deployment. Modern LLMs require substantial computational resources, large memory footprints, and high energy consumption, making them difficult to run on edge devices, real-time systems, or cost-constrained environments. Their inference latency can hinder interactive applications, and frequent retraining or fine-tuning exacerbates the computational burden. These limitations have created a strong need for efficient model compression techniques, such as pruning \cite{sun2023simple}, knowledge distillation \cite{gu2023minillm}, and quantization \cite{lang2024comprehensive}, which aim to reduce model size, improve inference speed, and lower resource requirements while maintaining accuracy. Motivated by these issues, this research explores how such optimization strategies can make large language models more scalable, accessible, and practical for real-world use. Quantization in deep learning is an optimization technique that reduces the precision of a model's numerical values, typically converting them from floating-point numbers to lower-precision integers. The authors in \cite{lang2024comprehensive, yao2023comprehensive} have published a comprehensive study on the quantization of LLMs range mapping viz, affine quantization, scale quantization, quantization techniques viz., post-training quantization, quantization-aware training, weight quantization and activation-aware weight quantization. Although post-training quantization techniques improve LLM's computational efficiency and memory footprint, their hand-crafted quantization settings result in poor performance, particularly in very low-bit quantization. This issue is addressed in omnidirectionally calibrated quantization i.e., OmniQuant technique for LLMs. To address the issue of substantial training resources in quantization aware training, effientQAT method was proposed consisting of Block-wise training of all parameters and end-to-end training of quantization parameters \cite{chen2025efficientqat}. Authors in \cite{li2023loftq} applied low rank adaptation along with quantization aware training simultaneously which reduces the difference between the full-precision and quantized models and greatly enhances generalization in subsequent challenges. Pruning in deep learning is a technique to reduce the size and complexity of a neural network by removing less important parameters, like weights, neurons, or entire layers. In one approach proposed, each weight matrix is parameterized using its low-rank factorization and adaptively rank-1 components are eliminated during training \cite{wang2020structured}.  A novel and effective pruning method titled Wanda (pruning by Weights and activations), designed to induce sparsity in pretrained LLMs \cite{sun2023simple}. A batched greedy pruning method names SlimGPT for rapid and near-optimal pruning is proposed which enhances the accuracy of head-wise pruning error estimation through grouped Cholesky decomposition \cite{ling2024slimgpt}. Authors in \cite{an2024fluctuation} proposed Fluctuation-based Adaptive Structured Pruning  which formulated structured importance metrics, adaptively searched the global compressed model, and implemented compensation mechanisms to mitigate performance loss.

While individual compression techniques—quantization \cite{dettmers2023qlora}, pruning \cite{ma2023llm}, and knowledge distillation \cite{hinton2015distilling}, have been extensively studied in isolation, real-world deployment scenarios often require combining multiple techniques to achieve aggressive compression ratios while maintaining acceptable performance \cite{zhu2024survey}. However, the existing literature provides limited guidance on the optimal ordering of these techniques when applied sequentially on small scale LLMs. Recent surveys \cite{wan2023efficient} identify this as a critical gap: different orderings may exhibit synergistic or antagonistic interactions, and certain sequences may be infeasible due to technical constraints (e.g., quantization's incompatibility with gradient-based training). The foundational "Deep Compression" work in \cite{han2016eie} used a pipeline of pruning, quantization, and encoding to drastically reduce the AlexNet model's size. This same principle of applying techniques in a sequence can be extended to modern small-scale LLMs. This work systematically explores compression technique orderings to identify optimal strategies for practitioners deploying compressed LLMs in resource-constrained environments.

This research paper is organized as follows: Section \ref{sec:methodology} describes the methodology explaining single strategy baseline compression techniques in section \ref{subsec:compression_tech}. This includes knowledge distilation in section \ref{subsubsec:knowledge_distillation}, pruning in section \ref{subsubsec:structured_pruning}, and quantization in section \ref{subsubsec:quantization}. We propose our compresseion ordering strategy in section \ref{subsec:compression_strategy} with specific focus on $6$ three-technique sequences explained in section \ref{subsubsec:three_technique_seq}. Section \ref{sec:results} discussed the results and the analysis part and section \ref{sec:conclusion} concludes the paper.


\section{Methodology}
\label{sec:methodology}
The methodology begins by outlining each individual compression strategy—knowledge distillation, pruning and quantization—highlighting how each technique independently reduces model size or computation while maintaining acceptable performance. However, relying on any single method alone introduces limitations, such as accuracy degradation from aggressive quantization, reduced representational capacity in distilled models, or structural instability caused by extensive pruning. To address these shortcomings, we propose a sequential three-stage pipeline that integrates the strengths of all three approaches. We leverage multiple combinations of knowledge distillation, pruning in various orders.
\subsection{Compression Techniques}
\label{subsec:compression_tech}
\subsubsection{Knowledge Distillation (KD)}
\label{subsubsec:knowledge_distillation}
Knowledge distillation is an effective model compression technique used to transfer the capabilities of a large, high-performing language model (teacher) into a smaller, more efficient model (student). In the context of LLMs, this process involves guiding the student model to mimic the teacher’s behavior by learning from its soft predictions, hidden representations, or intermediate reasoning patterns. Through this targeted supervision, the student model can capture much of the teacher’s linguistic and reasoning ability while requiring significantly fewer parameters and computational resources. 

In this paper, knowledge distillation was performed using Qwen2.5-7B \cite{team2024qwen2} as the teacher model to guide the training of compressed student models, following the framework established in \cite{hinton2015distilling}. The teacher model was loaded in 8-bit precision using BitsAndBytes quantization \cite{dettmers2022gpt3} to manage GPU memory constraints while maintaining high-quality probability distributions. 

\textbf{Loss Function}: The distillation loss combined two components following the dual-objective approach of \cite{hinton2015distilling}:
\begin{enumerate}
    \item Task Loss: Standard cross-entropy loss between student predictions and ground-truth labels
    \item Distillation Loss: KL divergence between teacher and student probability distributions. 
\end{enumerate}
The combined loss was formulated as:
\begin{equation}
\label{eq:KD_loss}
L_\text{total} = (1 - \alpha) × L_\text{task} + \alpha × L_\text{distill}
\end{equation}
where $\alpha = 0.3$ , giving $70\%$ weight to task performance and $30\%$ to teacher mimicry. This weighting scheme balances learning from both hard labels and soft targets, as recommended in the distillation literature \cite{sanh2019distilbert, jiao2020tinybert}

\textbf{Temperature Scaling}: Both teacher and student logits were scaled by temperature $T = 4.0$ before computing softmax distributions, following \cite{hinton2015distilling}. Temperature scaling smooths the probability distributions and reveals more information about the teacher's learned similarities between tokens \cite{guo2017calibration}. The distillation loss was scaled by $T^{2} = 16$ to compensate for the magnitude reduction caused by temperature scaling.

\textbf{Masking and Vocabulary Handling}: Attention masks were applied during KL divergence computation to ensure that padding tokens did not contribute to the loss, following standard practices in transformer training \cite{wolf2020transformers}. When teacher and student models had mismatched vocabulary sizes, logits were truncated to the minimum shared vocabulary size to enable proper distribution comparison.


\subsubsection{Structured Pruning (P)}
\label{subsubsec:structured_pruning}
Pruning reduces the size and computational cost of LLMs by removing weights or neurons that contribute minimally to the model’s predictions. By eliminating these less impactful parameters, pruning helps streamline the network, leading to faster inference and lower memory requirements while aiming to preserve overall performance. Structural pruning goes a step further by removing entire components—such as attention heads, neurons, or even blocks—rather than individual weights.

Structured pruning targeted the feed-forward network layers within each transformer block, specifically the intermediate dimension of the gate\_proj, up\_proj, and down\_proj layers, following the approach referenced in \cite{filters2016pruning} and recently adapted for LLMs in \cite{ma2023llm}. A pruning ratio of $30\%$ was applied uniformly across all layers, consistent with ratios shown to maintain model quality in recent pruning studies \cite{sun2023simple, kwon2022fast}.

Importance Score Calculation: Neuron importance was determined using a hybrid metric combining weight magnitude and activation statistics, inspired by the first-order Taylor expansion approach of \cite{molchanov2016pruning} and adapted for transformer architectures:
\begin{itemize}
    \item Weight Magnitude: Sum of absolute weights for each neuron \cite{han2015learning}.
\begin{equation}
I_{weight} = |W_{gate}[i, :]| + |W_{up}[i, :]|
\end{equation}
    \item Activation Magnitude: Average activation across calibration samples. $I_{activation} = E[|h_{i}|]$ over calibration batches
    \item Combined Score: $I_{combined} = 0.5 \times I_{\mathrm{weight norm}} + 0.5 \times I_{\mathrm{activation norm}}$
where normalization was performed by dividing by the maximum value within each layer to ensure equal scaling across different network depths \cite{ma2023llm}.
\end{itemize}


For each layer, neurons were ranked by combined importance score, and the bottom $30\%$ were removed. Structured pruning was performed by selecting top $70\%$ neurons (keep\_indices) based on importance, and reconstructing gate\_proj and up\_proj with reduced output dimensions, down\_proj with reduced input dimension to maintain dimensionality consistency.


\subsubsection{Quantization (Q)}
\label{subsubsec:quantization}
Quantization reduces the memory footprint and computational load of LLMs by representing weights and activations with lower-precision numerical formats instead of standard 16-bit or 32-bit floats. By compressing the model into fewer bits, quantization enables faster inference and more efficient deployment on GPUs, CPUs, and edge hardware. 

In this work, quantization is applied using the BitsAndBytes NF4 (4-bit NormalFloat) format, which preserves the statistical distribution of weights while significantly reducing storage requirements. NF4 is specifically designed for neural network weights that follow a normal distribution, providing information-theoretically optimal bin placement for minimal quantization error \cite{dettmers2023qlora}. This approach allows LLMs to maintain strong performance even under aggressive compression, making them more accessible for real-world, resource-limited environments. This approach builds upon earlier work in low-bit quantization \cite{jacob2018quantization} and has been shown to maintain model quality while achieving significant compression \cite{dettmers2023qlora}.



Inference-Only Constraint: BitsAndBytes 4-bit quantization produces inference-only models \cite{dettmers2023qlora}. The quantized weights are stored in a packed format that does not support gradient computation, similar to other post-training quantization methods \cite{frantar2022gptq}. This constraint is critical for determining valid compression orderings, as any technique requiring training (KD or post-pruning fine-tuning) must occur before final quantization.

\subsection{Compression Ordering Strategies}
\label{subsec:compression_strategy}
The baseline single technique models viz., KD \cite{hinton2015distilling}, P \cite{filters2016pruning}, Q \cite{dettmers2023qlora} show how these techniques individually help lower model complexity or computational load while still retaining reasonable accuracy yet they have their own drawbacks when used alone. These techniques suffer performance loss from heavy quantization, limited expressiveness in distilled models, or architectural imbalance due to excessive pruning. To overcome these issues, we present a unified three-step framework that combines the advantages of all three methods in a cohesive pipeline.We systematically evaluated different orderings of the three compression techniques viz., KD, P, Q to identify optimal strategies, addressing a gap in the literature where most works study these techniques in isolation \cite{zhu2024survey, wan2023efficient}. These distinct compression sequences were implemented and compared:


\subsubsection{Three-Technique Sequences (Primary Focus)}
\label{subsubsec:three_technique_seq}
\textbf{Direct Sequences} (no intermediate quantization):
\begin{itemize}
    \item KD-P-Q: Knowledge Distillation → Pruning → Quantization
    \item P-KD-Q: Pruning → Knowledge Distillation → Quantization
\end{itemize}

\textbf{Dequantization-Based Sequences} (quantization in middle or beginning):
\begin{itemize}
    \item Q-P-KD: Quantization → Dequantization → Pruning → Knowledge Distillation → Re-quantization 
    \item Q-KD-P: Quantization → Dequantization → Knowledge Distillation → Pruning → Re-quantization
    \item KD-Q-P: Knowledge Distillation → Quantization → Dequantization → Pruning → Re-quantization
    \item P-Q-KD: Pruning → Quantization → Dequantization → Knowledge Distillation → Re-quantization
\end{itemize}
Please note that the dequantization-based sequences involve an implicit dequantization step (D) that is not shown in the sequence notation for brevity. For example, Q-P-KD actually represents Q-D-P-KD-Q.


\subsection{Rationale for Compression Ordering}
\label{subsec:Rationale_for_Compression_Ordering}
The selection of compression orderings was guided by the following theoretical considerations:
\begin{enumerate}
    \item Quantization Last Principle: BitsAndBytes 4-bit quantization produces inference only models \cite{dettmers2023qlora}, making it incompatible with subsequent training based techniques. Therefore, all valid sequences must have \textbf{Q} as the final step, except for dequantization-based approaches. This constraint is common to most post-training quantization methods \cite{frantar2022gptq}.
    \item Dequantization Quality Concerns: Dequantization introduces information loss and potential weight corruption due to the irreversible nature of quantization \cite{nagel2021white}. Sequences with early quantization (Q-*) test the hypothesis that this quality degradation significantly impacts subsequent compression effectiveness.
    \item Knowledge Transfer Order: KD-P-Q tests whether early knowledge distillation provides a stronger foundation for pruning, while P-KD-Q tests whether pruning first creates a more efficient student for distillation. This addresses questions about the interaction between distillation and structural compression raised by recent work (\cite{sanh2019distilbert, ma2023llm}.
    \item Compression Synergy: Different orderings explore whether certain compression techniques enhance or interfere with subsequent techniques (e.g., does KD make pruning more effective by creating more robust representations?). Understanding these interactions is identified as a key challenge in the compression literature \cite{zhu2024survey, miao2025towards}.
\end{enumerate}
These ordering strategies allow comprehensive analysis of compression technique interactions and identification of optimal workflows for practical deployment scenarios, addressing a gap noted in recent surveys \cite{wan2023efficient}.

\section{Results and Analysis}
\label{sec:results}
\subsection{Experimental Setup}
\label{subsec:exp_setup}
\subsubsection{Hardware and Software Configuration}
\label{subsubsec:hard_soft_config}
All experiments were conducted on a single NVIDIA A6000 GPU with $48$ GB CUDA memory. The base model used for compression was Qwen2.5-3B \cite{team2024qwen2}, a state-of-the-art large language model with $3$ billion parameters. The experimental framework was implemented using PyTorch 2.0+ \cite{paszke2019pytorch} with the Transformers library v4.36+ \cite{wolf2020transformers} and BitsAndBytes v0.41+ \cite{dettmers2023qlora} for quantization operations. Training precision utilized BFloat16 (BF16) when supported by hardware, following the mixed precision training paradigm \cite{micikevicius2017mixed}, otherwise defaulting to FP$32$ for numerical stability.
\subsubsection{Dataset}
\label{subsubsec:dataset}
We utilized the HuggingFaceH4/ultrachat\_200k dataset \cite{ding2023enhancing} for all training, calibration, and some evaluation procedures. The dataset consists of multi-turn conversational data suitable for chat-based language model fine-tuning. For knowledge distillation experiments, 10$\%$ of the training split (approximately $10000$ samples) was used, following standard practices in neural network distillation \cite{sanh2019distilbert}. For structured pruning experiments, $2000$ samples were allocated, with $500$ samples dedicated to importance score calculation and $1500$ samples for post-pruning fine-tuning, consistent with calibration requirements in recent pruning literature. All text sequences were tokenized with a maximum length of $512$ tokens and padded to maintain uniform batch processing. For evaluation, we used the Stanford Question Answering Dataset (SQuAD), a reading comprehension dataset consisting of questions posed on Wikipedia articles \cite{rajpurkar-etal-2016-squad}. The dataset includes a. context selection selected: Full paragraphs from Wikipedia articles, b. Question formulation: Questions created by crowdworkers based on the paragraphs, and c. Answer extraction: Spans of text that answer the questions, annotated by humans. 

\subsection{Evaluation Metrics}
\label{subsec:evaluation_metrics}
The performance of the base LLM model and its compressed version using the single baseline compression techniques as well as the three-techniques sequence is evaluated using three state-of-the-art, and widely accepted metrics. All models were assessed on perplexity, model size (MB), and three quality metrics: G-Eval score, prompt alignment, perplexity and clarity. G-Eval is a framework that leverages an LLM-as-a-judge approach with chain-of-thought reasoning to assess model outputs using any user-defined criteria \cite{Ip_deepeval_2025}. In this study, correctness and relevance were adopted as core evaluation criteria to assess the factual accuracy and contextual alignment of candidate LLM-generated responses. Google's Gemini model is used as the judge model. Perplexity (PPL) is a common metric used to evaluate language models, and it measures how well a model predicts a sequence of words. A low perplexity score means the model is less confused. The Clarity metric is a composite of three sub-metrics that together viz., fluency, contextual coherence, and readability which quantify the linguistic and contextual quality of the LLM’s response. Fluency is measured using DistilGPT2 to calculate the perplexity of the response as shown in the eq. \ref{eq:fluency_score}. The range of the flunecy score is between $0$ and $1$, where higher values indicate greater fluency. Contextual coherence evaluates the cosine similarity between the two vector embeddings of the LLM's response and context using the sentence transformer. Readability is computed using the Flesch Reading Ease (FRE) score. It is a rule-based metric that evaluates sentence length and syllable count to quantify the ease of reading a text. Finally, the clarity is computed as the average of Fluency, Contextual Coherence, and Readability, ensuring equal weight across the sub-dimensions. The prompt alignment metric employs an LLM-as-a-judge approach to assess how well your LLM’s generated outputs adhere to the instructions defined in the prompt template.
\begin{equation}
    \label{eq:fluency_score}
    \displaystyle \text{Fluency Score} = 1 - \frac{1}{1+\text{log}_{2}\left( \text{PPL} \right)}
\end{equation}

\begin{table}[]
\centering
\caption{Comprehensive Evaluation Results}
\label{tab:comprehensive_results}
\begingroup
\fontsize{28}{28}\selectfont
\resizebox{\textwidth}{!}{%
\begin{tabular}{|c|c|c|c|c|c|c|}
\hline
\textbf{Model}      & \textbf{G-Eval} & \textbf{\begin{tabular}[c]{@{}c@{}}Prompt \\ Alignment\end{tabular}} & \textbf{Clarity} & \textbf{Size (MB)} & \textbf{Perplexity} & \textbf{\begin{tabular}[c]{@{}c@{}}Compression \\ Ratio\end{tabular}} \\ \hline
\textbf{Base Model} & 0.830           & 0.670                                                                & 0.537            & 5886.01            & 3.418               & $1.0 \times$                                                          \\ \hline
KD                  & 0.790           & 0.800                                                                & 0.535            & 5886.01            & 3.226               & $1.0 \times$                                                          \\ \hline
P                   & 0.650           & 0.330                                                                & 0.499            & 4492.98            & 5.086               & $1.31 \times$                                                         \\ \hline
Q                   & 0.540           & 0.560                                                                & 0.515            & 1959.44            & 3.955               & $3.00 \times$                                                         \\ \hline
\textbf{P-KD-Q}     & \textbf{0.733}  & \textbf{0.300}                                                       & \textbf{0.492}   & \textbf{1600.13}   & \textbf{5.048}      & \textbf{$3.68 \times$}                                                \\ \hline
KD-P-Q              & 0.644           & 0.480                                                                & 0.504            & 1600.13            & 5.553               & $3.68 \times$                                                         \\ \hline
P-Q-KD              & 0.610           & 0.366                                                                & 0.491            & 1600.13            & 5.597               & $3.68 \times$                                                         \\ \hline
KD-Q-P              & 0.146           & 0.133                                                                & 0.520            & 1600.13            & 53.366              & $3.68 \times$                                                         \\ \hline
Q-KD-P              & 0.080           & 0.000                                                                & 0.503            & 1600.13            & 24.069              & $3.68 \times$                                                         \\ \hline
Q-P-KD              & 0.060           & 0.000                                                                & 0.501            & 1600.13            & 34.494              & $3.68 \times$                                                         \\ \hline
\end{tabular}}
\endgroup
\end{table}
.
\subsection{Overview}
\label{subsec:overview}
Table \ref{tab:comprehensive_results} presents the comprehensive evaluation results across all compression strategies. We evaluated ten model configurations: the base Qwen2.5-7B model, three single-technique baselines (KD, P, Q), and six three-technique orderings. KD's ability to improve prompt alignment beyond the base model suggests that the distillation process with Qwen2.5-7B as teacher successfully transferred instruction-following capabilities, consistent with findings in DistilBERT. Quantization's superior size reduction with minimal perplexity degradation demonstrates its dominant compression technique in practice. All six three-technique sequences achieved identical compression ratios ($3.68\times$, $1600.13$ MB), as expected since they apply the same operations. However, performance varied dramatically across orderings, revealing critical ordering dependencies. Among the six three-technique sequences, the sequence P-KD-Q showed exemplary performance compared to others. This applies pruning first, recovers performance through knowledge distillation, and finalizes with quantization. The other sequences with early quantization sequence i.e., Q-P-KD, Q-KD-P, and KD-Q-P exhibited catastrophic performance collapse, with perplexities $5-15 \times$ higher than successful sequences. These failures stem from dequantization quality degradation. When models are quantized early, they must be dequantized to enable subsequent training operations. Following are the key insights analyzed from the results shown in table \ref{tab:comprehensive_results}:
\begin{itemize}
    \item Quantization dominates single techniques: Q provides $3\times$ compression with minimal quality loss (G-Eval: $0.540$), far superior to pruning's $1.31\times$ compression at $0.650$ G-Eval.
    \item Three-technique sequences enable aggressive compression: P-KD-Q achieves $3.68\times$ compression while maintaining $0.733$ G-Eval, outperforming all single techniques except KD (which provides no compression).
    \item Diminishing returns: Moving from $3\times$ (Q alone) to $3.68\times$ (P-KD-Q) provides only $0.68\times$ additional compression while adding significant complexity, suggesting limited practical benefit of adding pruning to quantization-based workflows.
\end{itemize}

\section{Conclusion}
\label{sec:conclusion}
This study presents a comprehensive analysis of how knowledge distillation, structured pruning, and quantization interact when applied sequentially to compress Qwen LLMs. Our results show that while each technique individually contributes to model efficiency, their performance is heavily dependent on the ordering. Quantization consistently offered the highest compression with acceptable quality loss, whereas pruning introduced structural sparsity at the expense of increased perplexity. Among all evaluated sequences, P-KD-Q emerged as the most effective, achieving substantial compression while maintaining high G-Eval and clarity scores. Sequences featuring early quantization exhibited severe degradation due to irreversible quantization noise affecting downstream training. These findings highlight the importance of ordering-aware design when combining compression strategies. Overall, the work provides practitioners with a reliable pipeline for optimizing LLMs under constrained computational budgets.

Future research can explore adaptive pruning ratios and mixed-precision quantization, and alternate quantinzation techniques to further enhance compression without compromising accuracy. Additionally, extending this analysis to multimodal LLMs and larger-scale Qwen variants may reveal broader generalization patterns across architectures.

\bibliographystyle{elsarticle-num} 
\bibliography{ref}

@article{shaheen2021applications,
  title={Applications of Artificial Intelligence (AI) in healthcare: A review},
  author={Shaheen, Mohammed Yousef},
  journal={ScienceOpen Preprints},
  year={2021},
  publisher={ScienceOpen}
}

@article{cao2022ai,
  title={Ai in finance: challenges, techniques, and opportunities},
  author={Cao, Longbing},
  journal={ACM Computing Surveys (CSUR)},
  volume={55},
  number={3},
  pages={1--38},
  year={2022},
  publisher={ACM New York, NY}
}

@article{plathottam2023review,
  title={A review of artificial intelligence applications in manufacturing operations},
  author={Plathottam, Siby Jose and Rzonca, Arin and Lakhnori, Rishi and Iloeje, Chukwunwike O},
  journal={Journal of Advanced Manufacturing and Processing},
  volume={5},
  number={3},
  pages={e10159},
  year={2023},
  publisher={Wiley Online Library}
}

@article{jozefowicz2016exploring,
  title={Exploring the limits of language modeling},
  author={Jozefowicz, Rafal and Vinyals, Oriol and Schuster, Mike and Shazeer, Noam and Wu, Yonghui},
  journal={arXiv preprint arXiv:1602.02410},
  year={2016}
}

@article{chowdhary2020natural,
  title={Natural language processing},
  author={Chowdhary, KR1442},
  journal={Fundamentals of artificial intelligence},
  pages={603--649},
  year={2020},
  publisher={Springer}
}

@article{vaswani2017attention,
  title={Attention is all you need},
  author={Vaswani, Ashish and Shazeer, Noam and Parmar, Niki and Uszkoreit, Jakob and Jones, Llion and Gomez, Aidan N and Kaiser, {\L}ukasz and Polosukhin, Illia},
  journal={Advances in neural information processing systems},
  volume={30},
  year={2017}
}

@article{gu2023minillm,
  title={Minillm: Knowledge distillation of large language models},
  author={Gu, Yuxian and Dong, Li and Wei, Furu and Huang, Minlie},
  journal={arXiv preprint arXiv:2306.08543},
  year={2023}
}

@article{sun2023simple,
  title={A simple and effective pruning approach for large language models},
  author={Sun, Mingjie and Liu, Zhuang and Bair, Anna and Kolter, J Zico},
  journal={arXiv preprint arXiv:2306.11695},
  year={2023}
}

@inproceedings{lang2024comprehensive,
  title={A comprehensive study on quantization techniques for large language models},
  author={Lang, Jiedong and Guo, Zhehao and Huang, Shuyu},
  booktitle={2024 4th International Conference on Artificial Intelligence, Robotics, and Communication (ICAIRC)},
  pages={224--231},
  year={2024},
  organization={IEEE}
}

@article{radford2018improving,
  title={Improving language understanding by generative pre-training},
  author={Radford, Alec and Narasimhan, Karthik and Salimans, Tim and Sutskever, Ilya and others},
  year={2018},
  publisher={San Francisco, CA, USA}
}

@article{radford2019language,
  title={Language models are unsupervised multitask learners},
  author={Radford, Alec and Wu, Jeffrey and Child, Rewon and Luan, David and Amodei, Dario and Sutskever, Ilya and others},
  journal={OpenAI blog},
  volume={1},
  number={8},
  pages={9},
  year={2019}
}

@article{brown2020language,
  title={Language models are few-shot learners},
  author={Brown, Tom and Mann, Benjamin and Ryder, Nick and Subbiah, Melanie and Kaplan, Jared D and Dhariwal, Prafulla and Neelakantan, Arvind and Shyam, Pranav and Sastry, Girish and Askell, Amanda and others},
  journal={Advances in neural information processing systems},
  volume={33},
  pages={1877--1901},
  year={2020}
}

@inproceedings{devlin2019bert,
  title={BERT: Pre-training of deep bidirectional transformers for language understanding},
  author={Devlin, Jacob and Chang, Ming-Wei and Lee, Kenton and Toutanova, Kristina},
  booktitle={Proceedings of the 2019 conference of the North American chapter of the association for computational linguistics: human language technologies, volume 1 (long and short papers)},
  pages={4171--4186},
  year={2019}
}

@article{raffel2020exploring,
  title={Exploring the limits of transfer learning with a unified text-to-text transformer},
  author={Raffel, Colin and Shazeer, Noam and Roberts, Adam and Lee, Katherine and Narang, Sharan and Matena, Michael and Zhou, Yanqi and Li, Wei and Liu, Peter J},
  journal={Journal of machine learning research},
  volume={21},
  number={140},
  pages={1--67},
  year={2020}
}

@article{touvron2023llama,
  title={Llama: Open and efficient foundation language models},
  author={Touvron, Hugo and Lavril, Thibaut and Izacard, Gautier and Martinet, Xavier and Lachaux, Marie-Anne and Lacroix, Timoth{\'e}e and Rozi{\`e}re, Baptiste and Goyal, Naman and Hambro, Eric and Azhar, Faisal and others},
  journal={arXiv preprint arXiv:2302.13971},
  year={2023}
}

@article{bai2022constitutional,
  title={Constitutional ai: Harmlessness from ai feedback},
  author={Bai, Yuntao and Kadavath, Saurav and Kundu, Sandipan and Askell, Amanda and Kernion, Jackson and Jones, Andy and Chen, Anna and Goldie, Anna and Mirhoseini, Azalia and McKinnon, Cameron and others},
  journal={arXiv preprint arXiv:2212.08073},
  year={2022}
}

@article{chowdhery2023palm,
  title={Palm: Scaling language modeling with pathways},
  author={Chowdhery, Aakanksha and Narang, Sharan and Devlin, Jacob and Bosma, Maarten and Mishra, Gaurav and Roberts, Adam and Barham, Paul and Chung, Hyung Won and Sutton, Charles and Gehrmann, Sebastian and others},
  journal={Journal of Machine Learning Research},
  volume={24},
  number={240},
  pages={1--113},
  year={2023}
}

@article{bai2023qwen,
  title={Qwen technical report},
  author={Bai, Jinze and Bai, Shuai and Chu, Yunfei and Cui, Zeyu and Dang, Kai and Deng, Xiaodong and Fan, Yang and Ge, Wenbin and Han, Yu and Huang, Fei and others},
  journal={arXiv preprint arXiv:2309.16609},
  year={2023}
}

@article{yao2023comprehensive,
  title={A comprehensive study on post-training quantization for large language models},
  author={Yao, Zhewei and Li, Cheng and Wu, Xiaoxia and Youn, Stephen and He, Yuxiong},
  journal={arXiv preprint arXiv:2303.08302},
  year={2023}
}

@inproceedings{chen2025efficientqat,
  title={Efficientqat: Efficient quantization-aware training for large language models},
  author={Chen, Mengzhao and Shao, Wenqi and Xu, Peng and Wang, Jiahao and Gao, Peng and Zhang, Kaipeng and Luo, Ping},
  booktitle={Proceedings of the 63rd Annual Meeting of the Association for Computational Linguistics (Volume 1: Long Papers)},
  pages={10081--10100},
  year={2025}
}

@article{li2023loftq,
  title={Loftq: Lora-fine-tuning-aware quantization for large language models},
  author={Li, Yixiao and Yu, Yifan and Liang, Chen and He, Pengcheng and Karampatziakis, Nikos and Chen, Weizhu and Zhao, Tuo},
  journal={arXiv preprint arXiv:2310.08659},
  year={2023}
}

@inproceedings{wang2020structured,
  title={Structured pruning of large language models},
  author={Wang, Ziheng and Wohlwend, Jeremy and Lei, Tao},
  booktitle={Proceedings of the 2020 conference on empirical methods in natural language processing (emnlp)},
  pages={6151--6162},
  year={2020}
}

@article{dettmers2023qlora,
  title={Qlora: Efficient finetuning of quantized llms},
  author={Dettmers, Tim and Pagnoni, Artidoro and Holtzman, Ari and Zettlemoyer, Luke},
  journal={arXiv preprint arXiv:2305.14314},
  year={2023}
}

@article{ma2023llm,
  title={Llm-pruner: On the structural pruning of large language models},
  author={Ma, Xinyin and Fang, Gongfan and Wang, Xinchao},
  journal={Advances in neural information processing systems},
  volume={36},
  pages={21702--21720},
  year={2023}
}

@article{hinton2015distilling,
  title={Distilling the knowledge in a neural network},
  author={Hinton, Geoffrey and Vinyals, Oriol and Dean, Jeff},
  journal={arXiv preprint arXiv:1503.02531},
  year={2015}
}

@article{zhu2024survey,
  title={A survey on model compression for large language models},
  author={Zhu, Xunyu and Li, Jian and Liu, Yong and Ma, Can and Wang, Weiping},
  journal={Transactions of the Association for Computational Linguistics},
  volume={12},
  pages={1556--1577},
  year={2024},
  publisher={MIT Press 255 Main Street, 9th Floor, Cambridge, Massachusetts 02142, USA~…}
}

@article{wan2023efficient,
  title={Efficient large language models: A survey},
  author={Wan, Zhongwei and Wang, Xin and Liu, Che and Alam, Samiul and Zheng, Yu and Liu, Jiachen and Qu, Zhongnan and Yan, Shen and Zhu, Yi and Zhang, Quanlu and others},
  journal={arXiv preprint arXiv:2312.03863},
  year={2023}
}

@article{han2016eie,
  title={EIE: Efficient inference engine on compressed deep neural network},
  author={Han, Song and Liu, Xingyu and Mao, Huizi and Pu, Jing and Pedram, Ardavan and Horowitz, Mark A and Dally, William J},
  journal={ACM SIGARCH Computer Architecture News},
  volume={44},
  number={3},
  pages={243--254},
  year={2016},
  publisher={ACM New York, NY, USA}
}

@article{team2024qwen2,
  title={Qwen2 technical report},
  author={Team, Qwen and others},
  journal={arXiv preprint arXiv:2407.10671},
  volume={2},
  number={3},
  year={2024}
}

@article{ling2024slimgpt,
  title={Slimgpt: Layer-wise structured pruning for large language models},
  author={Ling, Gui and Wang, Ziyang and Liu, Qingwen},
  journal={Advances in Neural Information Processing Systems},
  volume={37},
  pages={107112--107137},
  year={2024}
}

@inproceedings{an2024fluctuation,
  title={Fluctuation-based adaptive structured pruning for large language models},
  author={An, Yongqi and Zhao, Xu and Yu, Tao and Tang, Ming and Wang, Jinqiao},
  booktitle={Proceedings of the AAAI Conference on Artificial Intelligence},
  volume={38},
  number={10},
  pages={10865--10873},
  year={2024}
}

@inproceedings{guo2017calibration,
  title={On calibration of modern neural networks},
  author={Guo, Chuan and Pleiss, Geoff and Sun, Yu and Weinberger, Kilian Q},
  booktitle={International conference on machine learning},
  pages={1321--1330},
  year={2017},
  organization={PMLR}
}

@article{filters2016pruning,
  title={Pruning filters for efficient convnets},
  author={Filters’Importance, Determine},
  journal={arXiv preprint arXiv:1608.08710},
  year={2016}
}

@article{kwon2022fast,
  title={A fast post-training pruning framework for transformers},
  author={Kwon, Woosuk and Kim, Sehoon and Mahoney, Michael W and Hassoun, Joseph and Keutzer, Kurt and Gholami, Amir},
  journal={Advances in Neural Information Processing Systems},
  volume={35},
  pages={24101--24116},
  year={2022}
}

@article{han2015learning,
  title={Learning both weights and connections for efficient neural network},
  author={Han, Song and Pool, Jeff and Tran, John and Dally, William},
  journal={Advances in neural information processing systems},
  volume={28},
  year={2015}
}

@article{molchanov2016pruning,
  title={Pruning convolutional neural networks for resource efficient inference},
  author={Molchanov, Pavlo and Tyree, Stephen and Karras, Tero and Aila, Timo and Kautz, Jan},
  journal={arXiv preprint arXiv:1611.06440},
  year={2016}
}

@inproceedings{jacob2018quantization,
  title={Quantization and training of neural networks for efficient integer-arithmetic-only inference},
  author={Jacob, Benoit and Kligys, Skirmantas and Chen, Bo and Zhu, Menglong and Tang, Matthew and Howard, Andrew and Adam, Hartwig and Kalenichenko, Dmitry},
  booktitle={Proceedings of the IEEE conference on computer vision and pattern recognition},
  pages={2704--2713},
  year={2018}
}

@article{frantar2022gptq,
  title={Gptq: Accurate post-training quantization for generative pre-trained transformers},
  author={Frantar, Elias and Ashkboos, Saleh and Hoefler, Torsten and Alistarh, Dan},
  journal={arXiv preprint arXiv:2210.17323},
  year={2022}
}

@software{Ip_deepeval_2025,
author = {Ip, Jeffrey and Vongthongsri, Kritin},
license = {Apache-2.0},
month = oct,
title = {{deepeval}},
url = {https://github.com/confident-ai/deepeval},
version = {3.7.2},
year = {2025}
}

@inproceedings{rajpurkar-etal-2016-squad,
    title = "{SQ}u{AD}: 100,000+ Questions for Machine Comprehension of Text",
    author = "Rajpurkar, Pranav  and
      Zhang, Jian  and
      Lopyrev, Konstantin  and
      Liang, Percy",
    editor = "Su, Jian  and
      Duh, Kevin  and
      Carreras, Xavier",
    booktitle = "Proceedings of the 2016 Conference on Empirical Methods in Natural Language Processing",
    month = nov,
    year = "2016",
    address = "Austin, Texas",
    publisher = "Association for Computational Linguistics",
    url = "https://aclanthology.org/D16-1264",
    doi = "10.18653/v1/D16-1264",
    pages = "2383--2392",
    eprint={1606.05250},
    archivePrefix={arXiv},
    primaryClass={cs.CL},
}

@article{micikevicius2017mixed,
  title={Mixed precision training},
  author={Micikevicius, Paulius and Narang, Sharan and Alben, Jonah and Diamos, Gregory and Elsen, Erich and Garcia, David and Ginsburg, Boris and Houston, Michael and Kuchaiev, Oleksii and Venkatesh, Ganesh and others},
  journal={arXiv preprint arXiv:1710.03740},
  year={2017}
}

@article{paszke2019pytorch,
  title={Pytorch: An imperative style, high-performance deep learning library},
  author={Paszke, Adam and Gross, Sam and Massa, Francisco and Lerer, Adam and Bradbury, James and Chanan, Gregory and Killeen, Trevor and Lin, Zeming and Gimelshein, Natalia and Antiga, Luca and others},
  journal={Advances in neural information processing systems},
  volume={32},
  year={2019}
}

@inproceedings{wolf2020transformers,
  title={Transformers: State-of-the-art natural language processing},
  author={Wolf, Thomas and Debut, Lysandre and Sanh, Victor and Chaumond, Julien and Delangue, Clement and Moi, Anthony and Cistac, Pierric and Rault, Tim and Louf, Remi and Funtowicz, Morgan and others},
  booktitle={Proceedings of the 2020 conference on empirical methods in natural language processing: system demonstrations},
  pages={38--45},
  year={2020}
}

@article{sanh2019distilbert,
  title={DistilBERT, a distilled version of BERT: smaller, faster, cheaper and lighter},
  author={Sanh, Victor and Debut, Lysandre and Chaumond, Julien and Wolf, Thomas},
  journal={arXiv preprint arXiv:1910.01108},
  year={2019}
}

@inproceedings{jiao2020tinybert,
  title={Tinybert: Distilling bert for natural language understanding},
  author={Jiao, Xiaoqi and Yin, Yichun and Shang, Lifeng and Jiang, Xin and Chen, Xiao and Li, Linlin and Wang, Fang and Liu, Qun},
  booktitle={Findings of the association for computational linguistics: EMNLP 2020},
  pages={4163--4174},
  year={2020}
}

@article{nagel2021white,
  title={A white paper on neural network quantization},
  author={Nagel, Markus and Fournarakis, Marios and Amjad, Rana Ali and Bondarenko, Yelysei and Van Baalen, Mart and Blankevoort, Tijmen},
  journal={arXiv preprint arXiv:2106.08295},
  year={2021}
}

@article{miao2025towards,
  title={Towards efficient generative large language model serving: A survey from algorithms to systems},
  author={Miao, Xupeng and Oliaro, Gabriele and Zhang, Zhihao and Cheng, Xinhao and Jin, Hongyi and Chen, Tianqi and Jia, Zhihao},
  journal={ACM Computing Surveys},
  volume={58},
  number={1},
  pages={1--37},
  year={2025},
  publisher={ACM New York, NY}
}

@article{dettmers2022gpt3,
  title={Gpt3. int8 (): 8-bit matrix multiplication for transformers at scale},
  author={Dettmers, Tim and Lewis, Mike and Belkada, Younes and Zettlemoyer, Luke},
  journal={Advances in neural information processing systems},
  volume={35},
  pages={30318--30332},
  year={2022}
}

@misc{ding2023enhancing,
      title={Enhancing Chat Language Models by Scaling High-quality Instructional Conversations}, 
      author={Ning Ding and Yulin Chen and Bokai Xu and Yujia Qin and Zhi Zheng and Shengding Hu and Zhiyuan Liu and Maosong Sun and Bowen Zhou},
      year={2023},
      eprint={2305.14233},
      archivePrefix={arXiv},
      primaryClass={cs.CL}
}

\end{document}